\documentclass[10pt,twocolumn,letterpaper]{article}

\usepackage{cvpr}      

\usepackage{graphicx}
\usepackage{amsmath}
\usepackage{amssymb}
\usepackage{booktabs}

\usepackage{multirow}
\usepackage{balance}

\usepackage[accsupp]{axessibility}

\usepackage[pagebackref,breaklinks,colorlinks]{hyperref}

\usepackage[capitalize]{cleveref}
\crefname{section}{Sec.}{Secs.}
\Crefname{section}{Section}{Sections}
\Crefname{table}{Table}{Tables}
\crefname{table}{Tab.}{Tabs.}

\def\cvprPaperID{*****} 
\def\confName{CVPR}
\def\confYear{2023}

\begin{document}

\title{Adaptive Sparse Pairwise Loss for Object Re-Identification}

\author{Xiao Zhou$^1\footnotemark[1]$
\quad Yujie Zhong$^{2}\footnotemark[2]$
\quad Zhen Cheng$^1\footnotemark[2]$
\quad Fan Liang$^2$
\quad Lin Ma$^2$\\
$^1$Department of Automation, BNRist, Tsinghua University, Beijing 100084 
\quad $^2$ Meituan Inc.\\
{\tt\small zhouxiao17@mails.tsinghua.edu.cn, jaszhong@hotmail.com, zcheng@mail.tsinghua.edu.cn}
}

\maketitle

\renewcommand{\thefootnote}{\fnsymbol{footnote}}
\footnotetext[1]{Work done during internship at Meituan.}
\footnotetext[2]{Corresponding author.}

\begin{abstract}
Object re-identification (ReID) aims to find instances with the same identity as the given probe from a large gallery. 
Pairwise losses play an important role in training a strong ReID network.
Existing pairwise losses densely exploit each instance as an anchor and sample its triplets in a mini-batch. 
This dense sampling mechanism inevitably introduces positive pairs that share few visual similarities, which can be harmful to the training.
To address this problem, we propose a novel loss paradigm termed Sparse Pairwise (SP) loss that only leverages few appropriate pairs for each class in a mini-batch, 
and empirically demonstrate that it is sufficient for the ReID tasks. Based on the proposed loss framework, we propose an adaptive positive mining strategy that can dynamically adapt to diverse intra-class variations. Extensive experiments show that SP loss and its adaptive variant AdaSP loss outperform other pairwise losses, and achieve state-of-the-art performance across several ReID benchmarks. Code is available at \tt\href{https://github.com/Astaxanthin/AdaSP}{https://github.com/Astaxanthin/AdaSP}.
\end{abstract}

\section{Introduction}
\label{sec:intro}
Object re-identification (ReID) is one of the most important vision tasks in visual surveillance. It aims at associating person/vehicle images with the same identity captured by different cameras in diverse scenarios. Recently, with the prosperity of deep neural networks in computer vision community, ReID tasks have rapidly progressed towards more sophisticated feature extractors \cite{suh2018part, zhao2017deeply, zhao2021heterogeneous, qian2022unstructured, meng2020parsing, dosovitskiy2020image} and more elaborate losses \cite{shi2016embedding, yuan2020defense, zeng2020hierarchical, sun2020circle}. Benefiting from quantifying the semantic similarity/distance of two images, metric losses are widely employed with the identity loss to improve ranking precision in the ReID task \cite{luo2019bag}. 

\begin{figure}
    \centering
    \includegraphics[scale = 0.65]{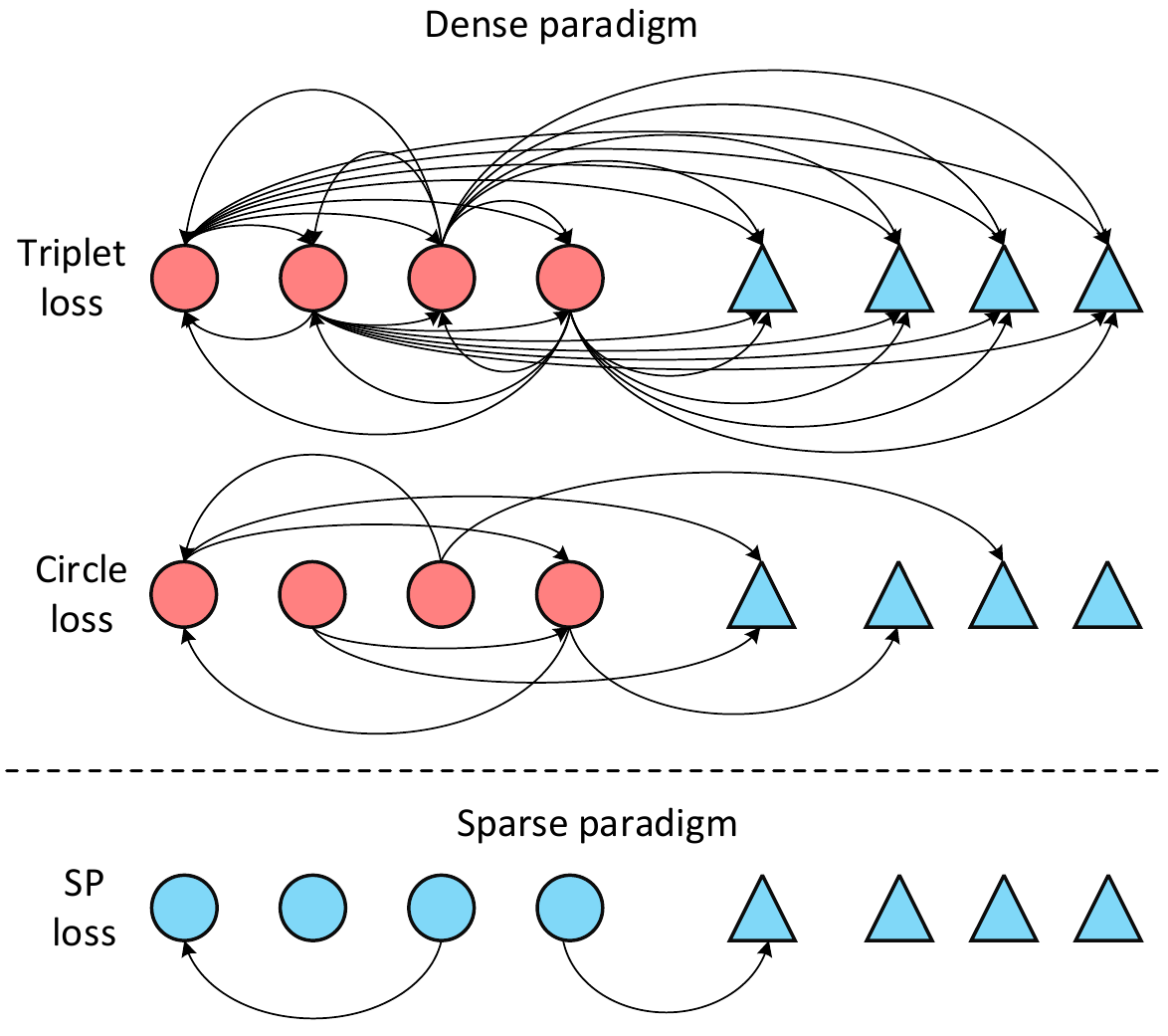}
    \caption{Difference between dense pairwise losses and our sparse pairwise (SP) loss. For simplicity, only two classes, marked by circles and triangles, are displayed. The red circles represent anchors. Dense pairwise losses, such as triplet loss\cite{hermans2017defense} and circle loss \cite{sun2020circle}, treat each instance as an anchor and mine its positive and negative pairs to construct a loss item. SP loss treats each class as a unit and separately excavates the most informative positive and negative pairs without using anchors.}
    \label{fig-1}
    \vspace{-2mm}
\end{figure}

\begin{figure}
    \centering
    \includegraphics[scale = 0.35]{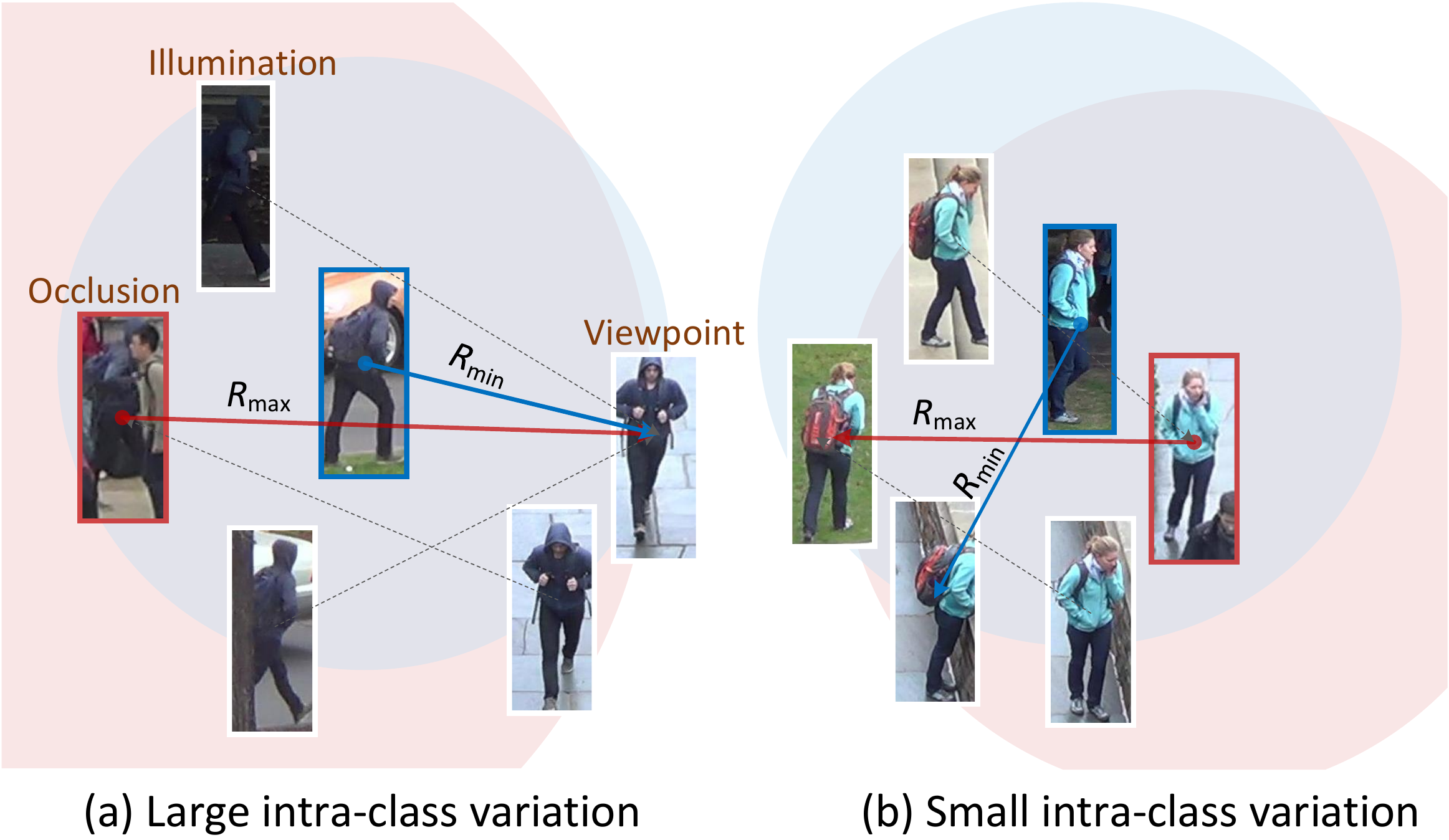}
    \caption{Different levels of intra-class variations in person ReID datasets. (a) Large intra-class variation caused by intensive occlusion, illumination changes, and different viewpoints. In this situation, mining the hardest positive pair, which forms the radius of the largest intra-class hypersphere, is harmful to metric learning, while the least-hard pair that shapes the radius of the smallest hypersphere is alternatively an appropriate one. (b) Small intra-class variation with highly similar visual features. The instances in both the hardest and the least-hard pairs share noticeable visual similarities in this case.}
    \label{fig0}
    \vspace{-2mm}
\end{figure}

Generally, metric losses serve for metric learning that aims to map raw signals into a low-dimensional embedding space where instances of intra-class are clustered and that of inter-class are separated. The pairwise framework, such as triplet loss \cite{hermans2017defense} and circle loss \cite{sun2020circle}, employs anchors to pull their positive pairs while pushing negatives apart. 
Most existing pairwise losses \cite{hermans2017defense, oh2016deep, harwood2017smart, sun2020circle} 
devote to exploiting all instances in a mini-batch, and densely anchoring each of them (which we term dense pairwise method) to sample its triplets. 
Although some hard sample mining methods~\cite{wang2019multi, wang2019ranked, sun2020circle, deng2022insclr} are developed to accelerate convergence or improve performance, these losses are still
computed in a dense manner (\ie top rows in Fig. \ref{fig-1}).
This dense design inevitably introduces harmful positive pairs \cite{shi2016embedding, bai2017scalable} that share few visual similarities and likely lead to bad local minima of optimization \cite{xuan2020hard} for metric learning, due to large intra-class variations (Fig. \ref{fig0}a) caused by illumination changes, occlusion, different viewpoints, etc. 

In this work, we anticipate that it is not necessary to employ all instances within a mini-batch since most of them are either trivial or harmful pairs. 
To this end, we propose a novel loss paradigm termed Sparse Pairwise (SP) loss that only leverages the most informative pairs for each class in a mini-batch without relying on dense instance-level anchors, as 
shown in Fig. \ref{fig-1} (bottom row). 
For the object ReID tasks, we empirically discover that using only few pairs of each class is sufficient for the loss computation as long as the appropriate ones are mined.
Based on the proposed loss framework, we conduct a rigorous investigation of the positive mining strategy.

Compared to negative mining which has been actively studied \cite{schroff2015facenet, harwood2017smart, xuan2020hard}, positive mining remains under-explored. Generally, an appropriate positive pair should be capable of adapting to different levels of intra-class variations. For example, an occluded pedestrian in a side viewpoint or dark illumination almost shares no visual similarities with his front 
perspective in Fig. \ref{fig0}a, while the instance (blue bounding box) in a clear illumination without any occlusions can bridge them, thus forming an appropriate pair for training. 
To obtain appropriate positive pairs, some works \cite{shi2016embedding, levi2021rethinking, xuan2020improved} attempt to excavate “moderate” or easy positive pairs, yet still rely on dense anchors. 

In this work, we first propose a least-hard positive mining strategy to address large intra-class variations. Inspired by a geometric insight, we find that the hardest positive pair in a class shapes the radius of the largest hypersphere covering all intra-class samples. Nevertheless, it can be excessively large and thus introduces overwhelming visual differences on account of large intra-class variations. In this case, the least-hard positive pair that constructs the smallest hypersphere can be utilized instead as a more appropriate one sharing noticeable
visual similarities (Fig. \ref{fig0}a). In addition, to handle classes with diverse intra-class variations (Fig. \ref{fig0}), we develop an adaptive mining approach that automatically reweights these two kinds of positive pairs, which shapes a dynamic intra-class hypersphere according to the particular situation of the class.
The main contributions of this paper are summarized as follows:

\begin{itemize}
    \item We propose a novel pairwise loss framework -- Sparse Pairwise loss -- that only leverages few informative positive/negative pairs for each class in a mini-batch.
    \item We propose a least-hard positive mining strategy to address large intra-class variations and further endow it with an adaptive mechanism according to different levels of intra-class variations in each mini-batch.
    \item The proposed AdaSP loss is evaluated on various person/vehicle ReID datasets and outperforms existing dense pairwise losses across different benchmarks. 
\end{itemize}

\section{Related Work}

\textbf{Object ReID.} Object ReID, including person and vehicle re-identification, is widely applied in intelligent surveillance systems across non-overlapping cameras \cite{chen2017person, khan2019survey}. Currently, feature representation learning and deep metric learning have become routine steps for ReID tasks by replacing handcrafted features \cite{gray2008viewpoint, matsukawa2016hierarchical} and distance metric learning \cite{koestinger2012large, xiong2014person, liao2015efficient} in early studies. To alleviate the misalignment in feature representations, many studies \cite{suh2018part, zhao2017deeply, cheng2016person, li2017learning, wang2018learning, zhao2021heterogeneous, qian2022unstructured, meng2020parsing} combine local aggregated features with global representation and achieve significant improvements in person/vehicle ReID tasks. Recently, the vision transformer \cite{dosovitskiy2020image, touvron2021training} has been introduced into ReID studies \cite{he2021transreid, qian2022unstructured, li2022dip} to obtain more discriminative feature representations. Besides, Fu et al. \cite{fu2021unsupervised, fu2022large} manage to perform pre-training on a large-scale unlabeled person ReID dataset to improve the generalization performance.

\textbf{Deep metric learning.} Metric learning mainly focuses on designing a metric loss that is capable of maximizing intro-similarities while minimizing inter-similarities in the embedding space. By treating each instance as an anchor, many pairwise metric losses, such as triplet loss \cite{hermans2017defense} and its variants \cite{zeng2020hierarchical, yuan2020defense, cheng2016person, ge2018deep}, quadruplet loss \cite{chen2017beyond}, and lifted structure loss \cite{oh2016deep}, demand the similarities of anchor-positive pairs to be higher than that of anchor-negatives. To further empower hard samples, multi-similarity loss \cite{wang2019multi}, rank listed loss \cite{wang2019ranked} and circle loss \cite{sun2020circle} reweight each pair according to its relative similarities and its optimal state, respectively. Beyond ReID and image retrieval tasks, Prannay et al. \cite{khosla2020supervised} propose a supervised contrastive loss to address image classification tasks. However, these pairwise losses highly rely on instance-level anchors and require densely mining positive pairs for each anchor. 

\textbf{Hard sample mining.} To alleviate the dominance of trivial triplets during the training, hard sample mining plays a critical role in metric learning. Alexander et al. \cite{hermans2017defense} empirically demonstrate that triplet loss with mining the hardest positive and negative pairs in a mini-batch outperforms that with sampling all triplets. However, the hardest pairs likely cause bad local minima of optimization \cite{schroff2015facenet, oh2016deep, yu2018correcting, xuan2020hard} when the anchor-positive pairs are far less similar than the anchor-negatives, especially for large intra-class variations. The solution to this problem is approximately concentrating on mining “less-hard” negatives/positives. For instance, on the one hand, Florian et al. \cite{schroff2015facenet} utilize semi-hard negative pairs whose similarities are lower than their anchor-positives, while Ben et al. \cite{harwood2017smart} and Bhavya et al. \cite{vasudeva2021loop} excavate smart hard negatives and optimal hard negatives to avoid choosing the excessively hard ones. On the other hand, Hailin et al. \cite{shi2016embedding} select moderate positive pairs to reduce the influence of large intra-class variations, while other studies \cite{xuan2020improved, levi2021rethinking} directly conduct easy positive mining for the triplet generation.

\section{Methodology}

\subsection{Sparse Pairwise Loss}
Given a mini-batch that consists of $K$ random classes and each with $N$ instances, we denote the normalized embedding vector of the $n$th instance in the $i$th class as $\mathbf{z}_{n}^i=\frac{\mathbf{w}_{n}^i}{\lVert \mathbf{w}_{n}^i\rVert} $. Correspondingly, the metric between any two instances can be described by their dot product similarity. Instead of treating each instance as an anchor, we propose a Sparse Pairwise (SP) loss to separately excavate a positive and a negative pair for each class. The SP loss for a mini-batch can be formulated as:
\begin{equation}\label{eq0}
    \mathcal{L}_{SP}=\frac{1}{K}\sum_i^K{\log \left( 1+e^{\frac{S_{i}^{-}-S_{i}^{+}}{\tau}} \right)}
\end{equation}
Where $\tau$ is a temperature parameter. $S_i^{-}$ denotes the similarity of the chosen negative pair for the $i$th class from its all $(K-1)N\times N$ negative pairs. In this work, we simply mine the hardest one since it describes the minimum distance between classes and represents the most informative pair across all negative pairs. Correspondingly, the soft version of the hardest negative similarity is adopted:
\begin{equation}\label{eq1}
    S_{i}^{-}\approx \tau \log \left( \sum_{n=1}^N{\sum_{j=1,j\ne i}^K{\sum_{m=1}^N{e^{\frac{\mathbf{z}_{n}^{i}\mathbf{z}_{m}^{j}}{\tau}}}}} \right) 
\end{equation}

In analogy, $S_i^{+}$ represents the appropriate positive similarity, which will be described in detail in Section 3.2 and 3.3. Superficially, the SP loss demands the hardest negative similarity of each class to be smaller than its appropriate positive similarity. While essentially, it treats each class as a unit and maximizes the difference between its inter-class and intra-class similarities in a sparse manner, which is different from other pairwise approaches that densely generate loss items with each corresponding to a single instance.

\begin{figure}
    \centering
    \includegraphics[scale = 0.3]{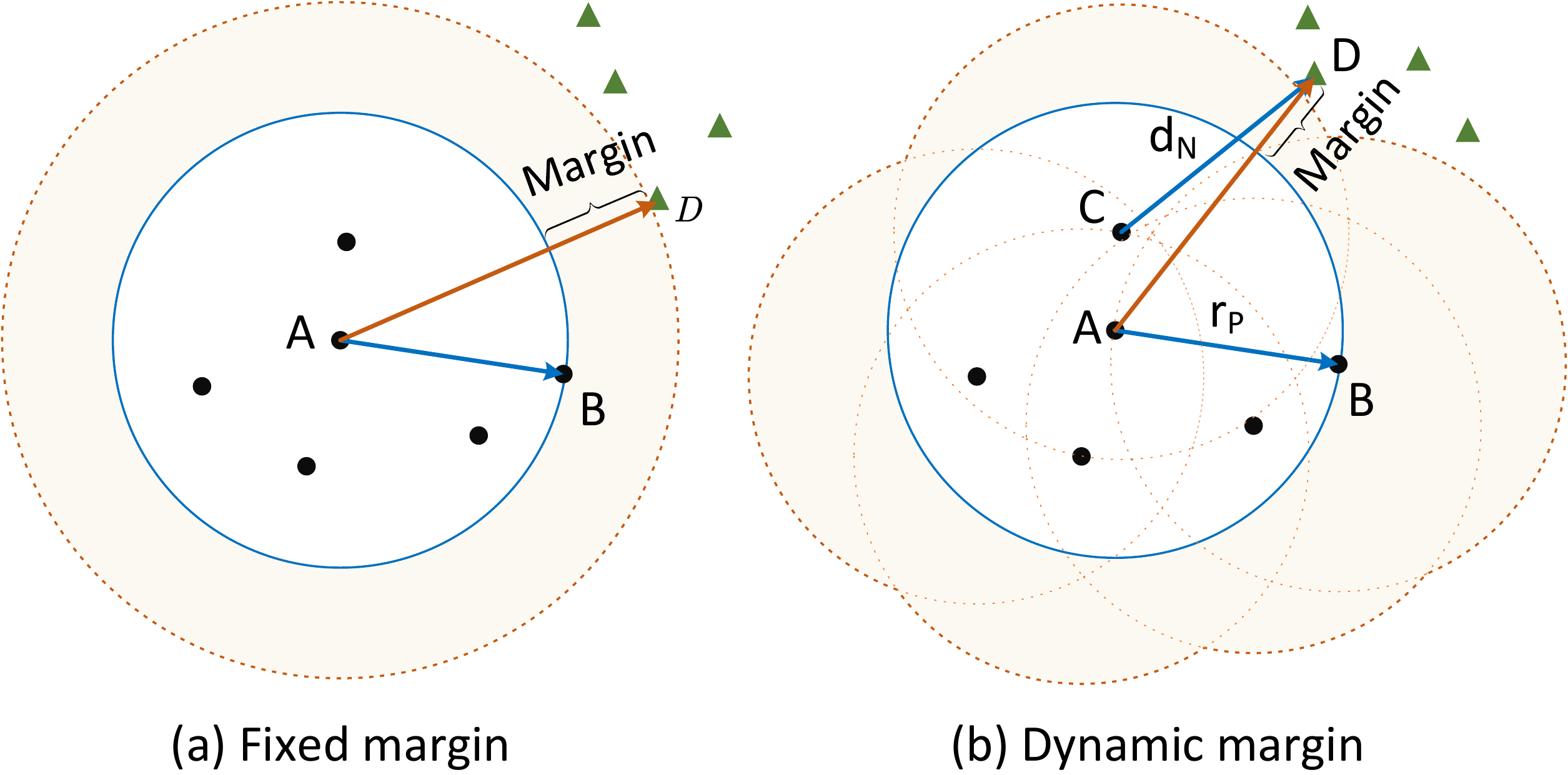}
    \caption{Schematic diagram for the difference of margin between SP loss and other dense pairwise losses. The black dots and green triangles suggest samples in different classes. The blue circles are centered at the anchor $A$ with a radius of its positive pair $AB$. The orange dash circles represent the border expanded by the margin. (a) Fixed margin adopted by dense pairwise losses. They require manually finetuning a hyper-parameter to enlarge the difference of similarities/distances between the positive and the negative pairs. (b) The dynamic margin for SP loss. Sparsely mining the hardest negative pair for each class rather than each instance shapes a margin that dynamically expands the class border and pushes the negative samples outside.}
    \label{fig1}
    \vspace{-2mm}
\end{figure}

\textbf{Implicit dynamic margin.} Another advantage of the proposed SP loss is that it does not require an explicit margin as a hyper-parameter which is broadly exploited in many dense pairwise losses \cite{sun2020circle, wang2019multi, hermans2017defense} to enlarge the difference between positives and negatives. Instead, an implicit dynamic margin is imposed by SP loss, as shown in Fig. \ref{fig1}b. Supposing the positive pair forms the radius, noted by $r_P$, of a hypersphere ($h_P$) that covers all intra-class samples, SP loss requires the minimum inter-class distance, $d_N$, to grow above the intra-class radius. As a result, according to the inequality $r_P<d_N<d_{AD}$, negative samples, geometrically, are excluded outside a dynamic border generated by the union of $K$ hyperspheres with the radius of $r_P$. Hereafter, if we treat the center of $h_P$ as an anchor like other dense losses, the margin between its positive and negative pairs varies from $0$ to $r_P$ due to $r_P<d_{AD}<2r_P$, while the dense pairwise losses usually configure a fixed pre-defined margin, as shown in Fig. \ref{fig1}a.

\subsection{Least-hard Positive Mining}
Generally, the similarity of the hardest positive pair in the $i$th class can be generated by a soft version:
\begin{equation}\label{eq2}
    S_{i,h}^{+}\approx -\tau \log \left( \sum_{n=1}^N{\sum_{m=1}^N{e^{-\frac{\mathbf{z}_{n}^{i}\mathbf{z}_{m}^{i}}{\tau}}}} \right) 
\end{equation}

\begin{figure}
    \centering
    \includegraphics[scale = 0.25]{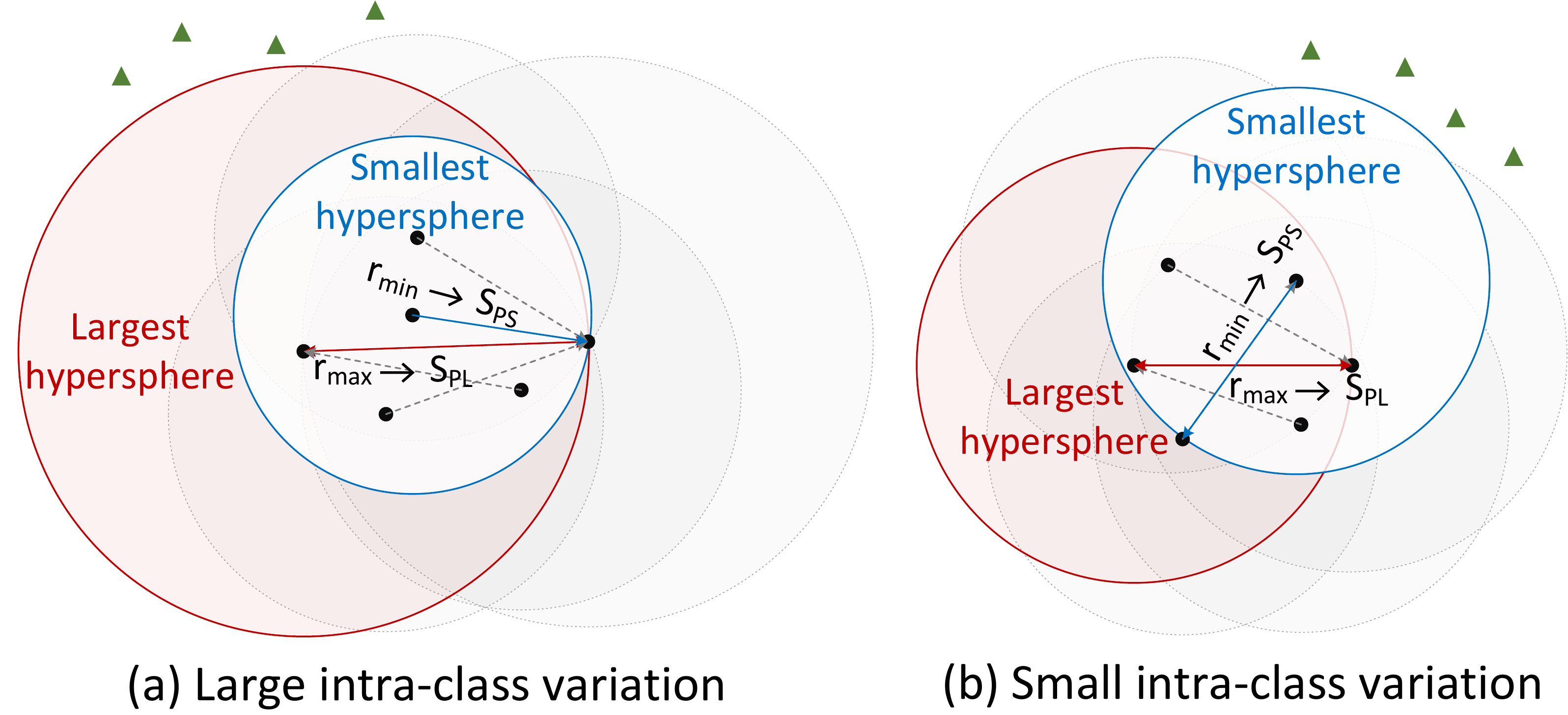}
    \caption{Schematic diagram for different levels of intra-class variations. The black dots and green triangles suggest samples in different classes. The hardest and the least-hard positive pairs shape the radii of the largest (red) and the smallest (blue) hypersphere that covers all intra-class samples, respectively. Gray circulars suggest hyperspheres centered at other instances with different radii corresponding to their farthest pairs. $S_{PL}$ and $S_{PS}$ denote the positive similarities corresponding to the largest and the smallest hyperspheres, respectively. (a) Large intra-class variation. A significant difference in size between the largest and the smallest hypersphere can be observed. (b) Small intra-class variation. In this situation, all hyperspheres share a similar size.}
    \label{fig2}
    \vspace{-2mm}
\end{figure}

Geometrically, the hardest positive pair shapes the radius of a hypersphere that is centered at either instance of the pair and covers all other instances within this class, as shown in Fig. \ref{fig2}a. Similarly, each instance paired with its farthest positive sample can shape an intra-class hypersphere with different radii. Correspondingly, the similarity of the $n$th positive pair can be given by:
\begin{equation}\label{eq3}
    S_{i,n}^{+}\approx -\tau \log \left( \sum_{m=1}^N{e^{-\frac{\mathbf{z}_{n}^{i}\mathbf{z}_{m}^{i}}{\tau}}} \right) 
\end{equation}

Therefore, $N$ hyperspheres with different sizes can be obtained from each class, and conspicuously, the one constructed by the hardest positive pair is the largest hypersphere (Fig. \ref{fig2}). However, intricate illumination, intensive occlusion, and different viewpoints usually cause a large intra-class variation, which inevitably leads to an excessive visual difference of positive pairs, especially for the hardest one. For this case, inflating the similarity of the hardest pair in the embedding space will largely contradict the original data features and thus mislead the feature learning.

To address this problem, we soften the hard mining strategy by selecting the least-hard positive pair that constructs the smallest hypersphere (Fig. \ref{fig2}a), and the similarity of the least-hard positive pair can be given by:
\begin{equation}\label{eq4}
    \begin{aligned}
    S_{i,lh}^{+}& \approx \tau \log \left( \sum_{n=1}^N{e^{\frac{S_{i,n}^{+}}{\tau}}} \right) \\
    &=\tau \log \left( \sum_{n=1}^N{\frac{1}{\sum_{m=1}^N{e^{-\frac{\mathbf{z}_{n}^{i}\mathbf{z}_{m}^{i}}{\tau}}}}} \right)
    \end{aligned}
\end{equation}

Noticeably, the least-hard positive pair is still a hard pair for the instance centered at the smallest hypersphere, while it produces a weaker learning signal than the hardest one across the entire class, which can effectively reduce the influence of large intra-class variations.

\textbf{Bad local minima of optimization.} Dense pairwise losses suffer from bad local minima of optimization during training in a dataset with large intra-class variation and low inter-class variations \cite{xuan2020hard}. Because in this situation, it is very likely that an anchor image is more similar to its hardest negative instances than its positives. We consider them harmful triplets. 
Therefore, anchors play an important role in introducing harmful triplets.
On the contrary, SP loss can largely reduce the possibility of forming harmful triplets due to the anchor-free design and least-hard mining strategy. 
Apart from quantitative results, we also provide a theoretical proof to demonstrate that the expected percentage of harmful positive pairs sampled by SP in a mini-batch is lower than that by the dense sampling approaches, which is shown in Supplementary Materials.

\subsection{Adaptive Positive Mining}
Mining the positive pair that constructs the smallest hypersphere can work well for the categories with large intra-class variations. While for the data with small intra-class variations, the hardest positive pair can contribute to an accelerated training process. Unfortunately, a fixed sampling strategy is unable to cope with both situations dynamically. To this end, we propose an adaptive weighting strategy that dynamically balances these two positive pairs according to different intra-class variations. However, it is infeasible to accurately quantify the intra-class variations from merely a mini-batch in the embedding space, since it varies according to batch sampling, data augmentation, and training stages.

In this work, we posit that the difference between the largest and the smallest intra-class hyperspheres can be employed to approximate the feature variations within a category. Concretely, for the early training stage, the pairwise similarities computed in the embedding space change fast and the largest hypersphere is excessively huge. For this case, it is less risky to mine the least-hard positive pair to conduct metric learning. While in the late training stage, the pairwise similarities are updated in a slow and robust manner. If the intra-class variation of a class is small, its largest and smallest hyperspheres are close in a small size, shown in Fig. \ref{fig2}b. In contrast, the difference between both intra-class hyperspheres will be relatively intensive if a large intra-class variation is observed. 

Therefore, to measure their difference in a robust training stage, we compute the harmonic mean of the hardest and the least-hard similarities for the $i$th class in a mini-batch, which is given by:
\begin{equation}\label{eq5}
    h_i= \frac{2S_{i,lh}^{+}S_{i,h}^{+}}{S_{i,lh}^{+}\ + S_{i,h}^{+}}
\end{equation}
Correspondingly, if a large gap is observed between the hardest and the least-hard positive similarities, the harmonic mean is close to the smaller one, \ie the hardest positive similarity. While if their similarities are close, the harmonic mean will approach their normal mean. 
Moreover, the harmonic mean increases as the growth of the hardest positive similarity, which can largely reveal the significance of the hardest positive pair. 

As a result, we employ the harmonic mean of similarities (without gradient backpropagation) as an adaptive weight to balance the hardest and the least-hard positive pairs for each class: 
\begin{equation} \label{eq6}
    \alpha_i =\begin{cases}
	h_i&		S_{i,h}^{+}\geqslant 0\\
	0&		S_{i,h}^{+}<0\\
\end{cases}
\end{equation}
By employing the above weight for positive pairs, the weighted positive similarity can be given by:
\begin{equation}\label{eq7}
    S_{i}^{+}=\alpha _{i}S_{i,h}^{+}+\left( 1-\alpha _{i} \right) S_{i,lh}^{+}
\end{equation}
Accordingly, the entire metric loss for a mini-batch in Eq. \ref{eq0} can be rewritten as:
\begin{equation}\label{eq8}
    \mathcal{L}_{SP}=\frac{1}{K}\sum_i^K{\log \left( 1+e^{\frac{S_{i}^{-}-\left( \alpha _{i}S_{i,h}^{+}+\left( 1-\alpha _{i} \right) S_{i,lh}^{+}\right)}{\tau}} \right)}
\end{equation}
From the geometric perspective, the weighted positive similarities $S_{i}^{+}$ indicate a dynamic hypersphere with a varying radius, which enables the loss to automatically adapt to different levels of intra-class variations.

\textbf{Comparison with existing methods.}
Dense pairwise losses are generated from anchor-positive and anchor-negative pairs. Most of them \cite{oh2016deep, wang2019ranked, khosla2020supervised} densely sample each positive pair and mine hard negative pairs to establish triplets for metric learning. Some recent studies have focused on mining informative positive pairs to accelerate training convergence, as shown in Tab. \ref{tab0}. The triplet \cite{hermans2017defense} and the circle loss \cite{sun2020circle} exploit the hardest anchor-positive pairs. To avoid bad local minima caused by large intra-class variations, the MS \cite{wang2019multi} and the MP loss \cite{shi2016embedding} select anchor-positive pairs according to the similarities of the hardest negative pairs. The EP \cite{xuan2020improved} loss directly adopts the easiest anchor-positive pair to construct a triplet. 
Instead of relying on instance-level anchors, the proposed SP loss mines informative positive pairs for each class. SP-H and SP-LH aim to mine the hardest and the least-hard positive pairs. AdaSP dynamically changes positive similarities according to the levels of intra-class variations.

\begin{table}
\centering
\small
\caption{Comparison of positive mining among different losses. Triplet-BH represents the triplet loss with batch hard mining \cite{hermans2017defense}. SP-H, SP-LH, and AdaSP suggest SP loss with the hardest positive mining, the least-hard positive mining, and adaptive positive mining, respectively. $S^+_a$ and $S^+_c$ denote positive similarities of anchors and classes, respectively. $S^-_a$ suggests anchor-negative similarity.}
 \resizebox{\linewidth}{!}{
\begin{tabular}{ll}
\toprule
\centering
\label{tab0}
Loss   & Positive Mining \\ \midrule
Triplet-BH \cite{hermans2017defense}   & $S_{a}^{+}=\underset{n}{\min} \,\,S_{a,n}^{+}$  \\
MS \cite{wang2019multi} & $S_{a}^{+}<\underset{n}{\max} \,\,S_{a,n}^{-}+\epsilon $ \\
Circle \cite{sun2020circle}  & $S_{a}^{+}=\underset{n}{\min} \,\,S_{a,n}^{+}$ \\
MP \cite{shi2016embedding} & $S_{a}^{+}=\underset{n}{\min}\left\{ S_{a,n}^{+}\,\,|S_{a,n}^{+}>\underset{m}{\max}\,\,S_{a,m}^{-}\,\, \right\} $\\
EP \cite{xuan2020improved}  & $S_{a}^{+}=\underset{n}{\max} \,\,S_{a,n}^{+}$ \\ \midrule
SP-H & $S_{c}^{+}=\underset{n,m}{\min} \,\,S_{n,m}^{+}$ \\
SP-LH &$S_{c}^{+}=\underset{n}{\max}\,\underset{m}{\min} \,\,S_{n,m}^{+}$ \\
AdaSP & $S_{c}^{+}=\alpha\, \underset{n,m}{\min}\,\,S_{n,m}^{+}+\left( 1-\alpha \right) \underset{n}{\max}\,\underset{m}{\min}\,\,S_{n,m}^{+}$  \\ \bottomrule  
               
\end{tabular}
}
\vspace{-2mm}
\end{table}

\section{Experiments}
To demonstrate the advantage of SP loss in object ReID tasks, we first compare the performance of single SP loss to that of other current metric losses, including triplet loss \cite{hermans2017defense}, multi-similarity (MS) loss \cite{wang2019multi}, circle loss \cite{sun2020circle}, supervised contrastive (SupCon) loss \cite{khosla2020supervised} and easy positive (EP) triplet loss \cite{xuan2020improved}, on three person ReID dataset (Market-1501 \cite{zheng2015scalable}, DukeMTMC-reID \cite{ristani2016performance}, MSMT17 \cite{wei2018person}) and two vehicle datasets (VeRi-776 \cite{liu2016deep}, VehicleID \cite{liu2016vehicleid}). Second, we explore the robustness of metric losses towards increasing intra-class variations. 
Afterwards, we adopt different network architectures to evaluate the improvement of ReID performance achieved by combining the proposed AdaSP with identity loss. In addition, we also investigate the effect of hyper-parameters of SP loss on the performance of ReID tasks. At last, we extensively evaluate AdaSP loss on seven person/vehicle benchmarks and compare them to state-of-the-art approaches.

\begin{table*}
\centering
\small
\caption{ReID performance comparison between different metric losses on MSMT17, Market-1501, DukeMTMC-reID, VeRi-776 and VehicleID datasets. SP-H, SP-LH, and AdaSP represent our SP loss with different positive mining types. Bold font and underlining suggest the best and the second-best performance, respectively.}
\begin{tabular}{lcccccccccccc}
\toprule
\label{tab2}
\multirow{2}{*}{Dataset} & \multicolumn{2}{c}{\multirow{2}{*}{MSMT17}} & \multicolumn{2}{c}{\multirow{2}{*}{Market-1501}} & \multicolumn{2}{c}{\multirow{2}{*}{DukeMTMC}} & \multicolumn{2}{c}{\multirow{2}{*}{VeRi-776}} & \multicolumn{4}{c}{VehicleID}                                                      \\
                         & \multicolumn{2}{l}{} & \multicolumn{2}{c}{}
                         & \multicolumn{2}{c}{} & \multicolumn{2}{c}{}
                         & \multicolumn{2}{c}{Small} & \multicolumn{2}{c}{Medium} \\  
Loss    & mAP    & R1  & mAP    & R1  & mAP    & R1   & mAP  & R1  & R1  & R5  & R1   & R5          \\  \midrule
Triplet-BH \cite{hermans2017defense}    &  57.4  & 79.6  &79.4	&\textbf{89.6}	&86.3	&93.5 & \underline{77.2} &94.3  & 85.8 &97.4  & 80.4  & \textbf{95.9}           \\
MS \cite{wang2019multi}          &  45.4     & 69.5  &70.6	&82.8	&78.9 &90.6 	&73.0 &92.5  & 82.5   &  95.6  & 80.2   & 94.3           \\
Circle \cite{sun2020circle}    &  56.1     & 78.6  &78.7	&88.8	&86.0	&93.7 &72.9 &92.7  & 84.5   &  96.7  & 81.1   & \underline{95.8}           \\
SupCon \cite{khosla2020supervised}   &  38.2     & 61.9   &64.5	&79.7	&73.5	&86.2  &70.3  &89.6  & 79.7   &  97.1  & 73.8   & 94.0     \\  
EP \cite{xuan2020improved}  &  41.1    & 67.8  &67.4	&82.0	&76.0	&90.4  &57.2 &93.2  & 74.5    & 87.2   & 71.2  & 82.4       \\ \midrule
SP-H      & \textbf{61.0}     & \underline{82.0} &\textbf{80.5}	&\textbf{89.6}	&\textbf{87.5}	&\textbf{94.3}  &75.3  &93.0   &85.6	&\textbf{98.4}	&80.0	&95.4  \\
SP-LH      & 58.4     & 80.8  &77.0	  &87.7	  &83.8	  &93.4  &\underline{77.2}  &\underline{94.6}  &\textbf{87.0}	  &97.9	  &\textbf{81.9}	  &95.7           \\
AdaSP      & \underline{60.7}     & \textbf{82.3}  &\underline{80.1}	&\underline{89.5}	&\underline{86.8}	&\underline{94.1}  &\textbf{77.6}  &\textbf{94.8}  &\underline{86.4}	&\underline{98.3}	&\underline{81.8}	&\textbf{95.9}  \\ \bottomrule
\end{tabular}
\vspace{-2mm}
\end{table*}

\vspace{-3mm}
\paragraph{Datasets and evaluation metrics.}
Four person ReID datasets, including CUHK03 \cite{li2014deepreid}, Market-1501 \cite{zheng2015scalable}, DukeMTMC-reID \cite{ristani2016performance}, MSMT17 \cite{wei2018person}, and three vehicle ReID datasets, including VeRi-776 \cite{liu2016deep}, VehicleID \cite{liu2016vehicleid}, VERI-WILD \cite{lou2019veri} are employed to evaluate our method. The details of each dataset are summarized in Tab. S1 (Supplementary Materials).
We employ the widely adopted metrics including mean Average Precision (mAP), Rank1 (CMC@1), and Rank5 (CMC@5) to quantify ReID results. 

\vspace{-3mm}
\paragraph{Implementation details.}
For person ReID tasks, we adopt the popular implementation of MGN \cite{wang2018learning} with instance batch normalization (IBN) \cite{pan2018two} in Fast-ReID~\cite{he2020fastreid}
as the backbone unless specified. The size of the embedding feature is set to $256$. The input image size is set to $384\times 128$ or $256\times 128$ in different experiments.
For vehicle ReID tasks, we adopt ResNet-50~\cite{he2016deep} with IBN as the backbone. The size of the embedding feature is set to $2048$. The input image size is set to $256\times 256$ for all experiments.
For both tasks, the overall loss consists of an identity loss (\ie cross-entropy loss) and a metric loss (SP loss for our approach):
\begin{equation}\label{eq9}
    \mathcal{L} = \mathcal{L}_{ID} + \lambda \mathcal{L}_{SP}
\end{equation}
Adam optimizer \cite{kingma2014adam} with a weight decay factor of $5$e-$4$ and a warmup strategy with the base learning rate of $3.5$e-$4$ are adopted. We train the network for $60$ epochs on all datasets apart from the largest VERI-WILD ($120$ epochs). The batch size is set to $128$ for VeRi-776 as well as all person ReID datasets and $512$ for VehicleID/VERI-WILD.

\subsection{Comparison with Other Pairwise Losses}

In this section, we first evaluate and compare SP loss with other metric losses on MSMT17, Market-1501, DukeMTMC-reID, VeRi-776, and VehicleID datasets. To achieve a fair comparison, we keep the same experimental settings apart from the metric loss. The number of instances is first set to $8$ for all datasets. The size of input person images is set to $384\times128$. The performance comparison is shown in Tab. \ref{tab2}. It can be seen that SupCon loss achieves very limited performance on all datasets, as it samples all positive pairs for each anchor, which not only fails to strengthen informative pairs but also likely introduces harmful triplets. Although EP loss mines the easiest positives for each anchor, which can avoid introducing harmful triplets, it only achieves an mAP of $41.1$ on MSMT because it lacks the enhancement of hard but not harmful positive pairs. By mining less-hard positives and negatives, MS loss outperforms both SupCon and EP losses, while still falling behind the best methods. In addition, the triplet-BH loss and the circle loss achieve comparable performance on person ReID datasets because they essentially share the same mining strategy, \ie the hardest positives/negatives for each anchor. 
For SP loss, it can be observed that SP-H loss outperforms other dense pairwise losses on almost all datasets apart from VeRi-776, suggesting that sparse pairs are sufficient for the learning. In addition, SP-H outperforms SP-LH on the three person ReID datasets while achieving lower R1 than SP-LH on the vehicle ReID datasets. This may be because the majority of classes in person and vehicle datasets contain small and large intra-class variations, respectively, in the situation with 8 instances per identity.
Overall, AdaSP achieves the best or the second-best performance across all datasets since it leverages dynamic positive pairs to adapt to different levels of intra-class variations.

\subsection{Ablation Study}

\begin{table*}
\centering
\small
\caption{Experimental results on person ReID datasets using different backbones. The loss function consists of a cross-entropy loss and a metric loss (AdaSP for our approach, Triplet-BH \cite{hermans2017defense} for comparison). $^*$ denotes results copied from TransReID \cite{he2021transreid}.}
\begin{tabular}{llcccccccccc}
\toprule
\label{tab3}
\multirow{2}{*}{Backbone} & \multirow{2}{*}{Loss} & \multicolumn{2}{c}{MSMT17} & \multicolumn{2}{c}{Market1501} & \multicolumn{2}{c}{DukeMTMC} & \multicolumn{2}{c}{CUHK03-L} & \multicolumn{2}{c}{CUHK03-D} \\  
   &   & mAP   & R1    & mAP  & R1      & mAP    & R1  &mAP &R1 &mAP &R1  \\ \midrule

\multirow{2}{*}{ResNet-50} &Triplet-BH  & 52.6  & 77.1  & 81.8 & 92.6  & 73.9 & 86.1  & 64.1 & 66.1 & 60.9 & 63.5  \\
& AdaSP       &\textbf{56.9}  & \textbf{80.7} & \textbf{86.4} & \textbf{95.1} & \textbf{78.6}  & \textbf{89.0}  &  \textbf{67.1} & \textbf{69.1} & \textbf{65.0} & \textbf{67.8}  \\ \midrule

\multirow{2}{*}{ResNet-101} &Triplet-BH  &56.3	&79.2	&84.2	&93.2	&76.6  	&87.2	&66.9	&69.2	&63.2	&65.4  \\
& AdaSP       &\textbf{59.5}  & \textbf{82.1} & \textbf{87.6} & \textbf{95.0} & \textbf{79.6}  & \textbf{89.8}  &  \textbf{69.6} & \textbf{71.2} & \textbf{67.7} & \textbf{69.2}  \\ \midrule

\multirow{2}{*}{ViT-base} &Triplet-BH  &61.0$^*$	&81.8$^*$	&86.8$^*$  &\textbf{94.7}$^*$ &79.3$^*$  &88.8$^*$  &72.4	&74.5	&69.9	&71.6  \\

&AdaSP       & \textbf{62.2}  & \textbf{82.0}  & \textbf{87.0} & 94.3  & \textbf{80.0}  & \textbf{89.0} & \textbf{77.4} & \textbf{79.4} & \textbf{75.3} & \textbf{77.5} \\  \midrule

\multirow{2}{*}{DeiT-base} &Triplet-BH  &\textbf{61.4}$^*$  &81.9$^*$	&86.6$^*$	&94.4$^*$  &78.9$^*$  &89.3$^*$	&70.9	&72.1	&67.8	&70.2  \\

&AdaSP       &61.3   &\textbf{82.5}   & \textbf{87.0} & \textbf{94.5}  & \textbf{79.6}  & \textbf{89.5} & \textbf{76.7} & \textbf{78.7} & \textbf{73.5} & \textbf{75.7} \\

\bottomrule
\end{tabular}
\vspace{-2mm}
\end{table*}

\begin{figure}
    \centering
    \includegraphics[scale = 0.6]{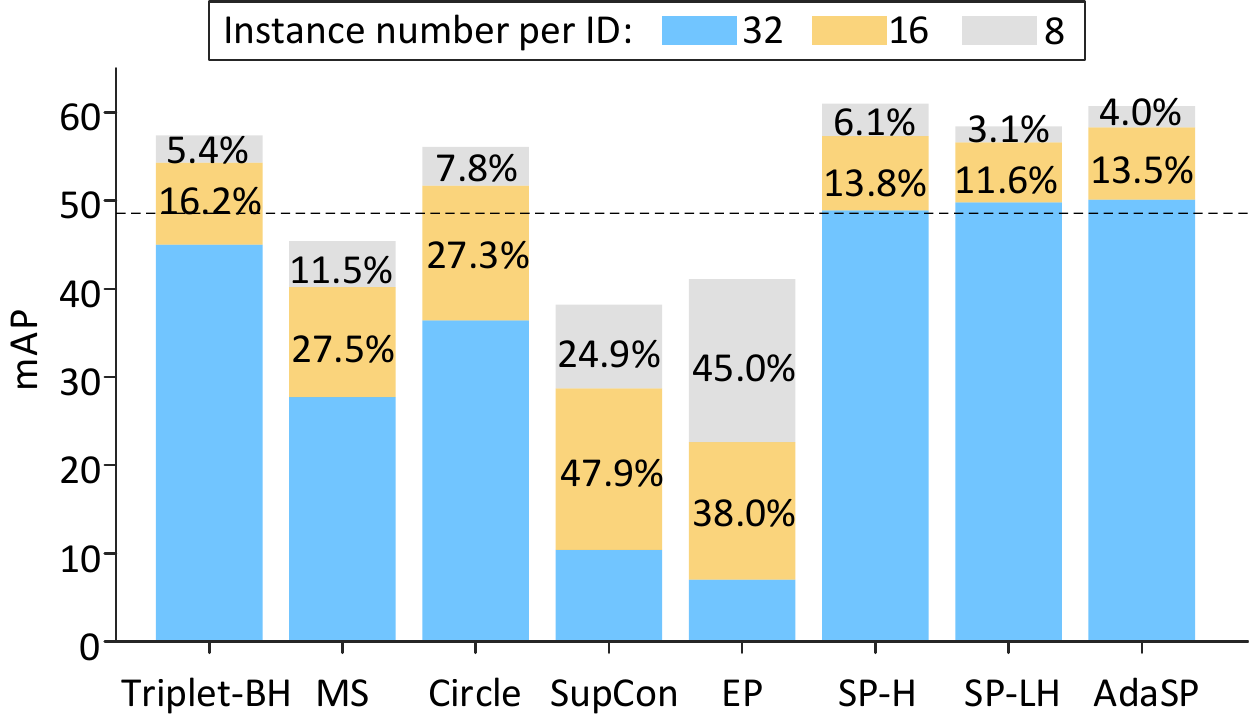}
    \caption{Comparison on the robustness of the performance regarding the number of instances per identity (including 8, 16, and 32) in a mini-batch on MSMT17 dataset. The numbers in each bar represent the percentage of dropped performance relative to the best performance achieved in 8 instances. The dashed line denotes the mAP performance of SP-H with 32 instances per identity.}
    \label{fig4}
    \vspace{-2mm}
\end{figure}

\paragraph{Robustness to increasing intra-class variations.}
To show the advantages of SP loss on large intra-class variations, we further increase the number of instances per identity from $8$ to $16$ and $32$, respectively. The experimental results are exhibited in Fig. \ref{fig4}. It can be seen that all metric losses suffer from a performance decline after increasing the instance number per identity in a mini-batch. This is mainly because the intra-class variations become intensive as the number of samples in a class rises. However, a significant difference in sensitivity towards the rising intra-class variations is observed among different metric losses. Specifically, the mAP performance of SupCon and EP losses dropped by more than $70\%$ after increasing the instance number to $32$. The circle and MS losses also lost more than $30\%$ mAP performance. The triplet loss with batch hard mining seems much less insensitive towards the rising intra-class variations than other losses, which, however, still performs worse than the three variants of SP loss. This indicates that our SP loss, even the hardest version SP-H, is less sensitive and more robust to large intra-class variations than other dense pairwise losses.

Moreover, it can be observed that when $8$ instances per identity are sampled, SP-H achieves the best performance, while after increasing the instance number, the minimal performance decline is achieved by SP-LH. This phenomenon suggests that the least-hard positive mining can substantially reduce the sensitivity towards increasing intra-class variations. Moreover, AdaSP draws the advantages from both SP-H and SP-LH, thus achieving the best mAP performance after enlarging the intra-class variations. Similar results can be concluded on the VeRi-776 dataset, shown in Fig. S1 in the Supplementary Materials.

\vspace{-3mm}
\paragraph{Generalizing to different networks.}
In order to demonstrate the advantage of SP loss on improving ReID performance when combined with identity loss, we replace the triplet loss with AdaSP and implement an extensive evaluation using different networks on four person ReID datasets. The size of input images is set to $256\times128$. The experimental results are exhibited in Tab. \ref{tab3}. 
Combined with the cross-entropy loss, AdaSP outperforms triplet loss on almost all datasets and all networks. Especially for ResNet-50 and ResNet-101, more than $3\%$ of mAP improvement is observed on each dataset. Additional results on ResNet50-IBN, ResNet-152 and MGN \cite{wang2018learning} are exhibited in Tab. S2 in Supplementary Material.
These encouraging experimental results suggest that the proposed AdaSP loss is capable of generating discriminative feature representations for pedestrian instances across different network architectures and can replace triplet loss to effectively work with cross-entropy loss.

\begin{table}
\centering
\footnotesize
\caption{Experimental results of the ablation study on the weight for metric loss $\lambda$ in MSMT17 dataset.}
\label{tab5}
\begin{tabular}{lccccccc}
\toprule
$\lambda$   & 0.01   &0.05    &0.1        & 0.2       & 0.3       & 0.4   & 0.5 \\ \midrule
mAP       & 66.0     & 66.8   & \textbf{67.1}   & 67.0   & 66.6     & 65.8    & 65.6   \\
R1        & 84.9     & 85.3   & \textbf{85.5}   & 85.4   & 85.0     & 84.9  & 84.9   \\   \bottomrule   
\end{tabular}
\vspace{-2mm}
\end{table}

\begin{table}
\centering
\caption{Experimental results of the ablation study on the temperature $\tau$ in MSMT17 dataset.}
\label{tab6}
\resizebox{\linewidth}{!}{
\begin{tabular}{lccccccccc}
\toprule
$\tau$ &0.01   & 0.02 & 0.03 &0.04  &0.05  &0.06  & 0.07  & 0.08  &0.09 \\ \midrule
mAP     &63.3  & 66.1 &66.9  & \textbf{67.1} &66.6  & 66.0 &65.8 &65.0 &64.6 \\
R1      &83.4  & 85.4 &85.4  & \textbf{85.5} &84.7  & 84.7 &84.5 &84.0 &83.2   \\   \bottomrule   
\end{tabular}
}
\vspace{-2mm}
\end{table}

\begin{table*}
\centering
\caption{Performance comparison with state-of-the-art approaches on person ReID datasets. AdaSP denotes our loss trained on MGN with IBN. $^*$ suggests the performance implemented with $160$ epochs ($100$ epochs more than our approach). Bold fonts and underlining suggest the best and the second-best performance, respectively. $-$ suggests the corresponding performance is not provided in the original study.}
 \resizebox{\linewidth}{!}{
\begin{tabular}{llccccccccccc}
\toprule
\label{tab7}
\multirow{2}{*}{Method} & \multirow{2}{*}{Backbone}  & \multirow{2}{*}{Input   size} & \multicolumn{2}{c}{MSMT17} & \multicolumn{2}{c}{Market1501} & \multicolumn{2}{c}{DukeMTMC} & \multicolumn{2}{c}{CUHK03-L} & \multicolumn{2}{c}{CUHK03-D} \\  
            &  &     & mAP   & R1    & mAP  & R1      & mAP    & R1  &mAP &R1 &mAP &R1  \\ \midrule
MGN \cite{wang2018learning}&  ResNet-50 & 384$\times$128  & --  & --  & 86.9 &\underline{95.7}  & 78.4 &88.7 & 67.4 & 68.0 & 66.0 &66.8  \\
OSNet \cite{zhou2019omni}&  OSNet & 256$\times$128  & 52.9  & 78.7  & 84.9 &94.8  & 73.5 &88.6 & -- & -- & 67.8 &72.3  \\
BAT-net \cite{fang2019bilinear} & GoogLeNet & 256$\times$128  &56.8  &79.5  & 87.4 &95.1  &77.3 &87.7 & 76.1 & 78.6 & 73.2 & 76.2  \\
ABD-Net \cite{chen2019abd} & ResNet-50 & 384$\times$128 & 60.8 & 82.3 & 88.3 & 95.6 & 78.6 & 89.0 & -- & -- & -- & -- \\
RGA-SC \cite{zhang2020relation} &ResNet-50 & 256$\times$128 & 57.5  &80.3  & 88.4  &\textbf{96.1} & -- & -- &77.4 &81.1 &74.5 &79.6  \\
ISP \cite{zhu2020identity} & HRNet-W32 & 256$\times$128 & -- & -- & 88.6  &95.3 & 80.0  & 89.6  & 74.1 & 76.5 & 71.4 & 75.2  \\
CDNet \cite{li2021combined} & CDNet & 256$\times$128 & 54.7 & 78.9 & 86.0  &95.1 & 76.8  & 88.6  &-- &-- &-- &--  \\
Nformer $^*$ \cite{wang2022nformer}& ResNet-50 & 256$\times$128  & 59.8 & 77.3 & \textbf{91.1} & 94.7 & \textbf{83.5} & 89.4  &78.0 &77.2  &74.7 &77.3 \\ \midrule
AdaSP & ResNet-50  &256$\times$ 128    & \underline{64.7}  & \underline{84.3}  & 89.0  & 95.1  & 81.5  & \underline{90.6}  & \underline{80.5} & \underline{82.1}  & \underline{78.0} & \underline{80.2}\\ 
AdaSP & ResNet-50   &384$\times$ 128     & \textbf{67.1}  & \textbf{85.5}   & \underline{89.8}   & 95.5  & \underline{83.0}  & \textbf{91.7} & \textbf{82.4} & \textbf{84.6}  & \textbf{80.1} & \textbf{82.0}\\ 
\bottomrule
\end{tabular}
}
\end{table*}

\begin{table*}
\centering
\caption{Performance comparison with state-of-the-art approaches on vehicle ReID datasets. Bold fonts and underlining suggest the best and the second-best performance, respectively. $^*$ suggests the reproduced performance under the same number of epochs as our experimental settings. $-$ suggests the corresponding performance is not provided in the original study. All backbones are ResNet-50.}
\label{tab8}
\resizebox{\linewidth}{!}{
\begin{tabular}{lcccccccccccccc}
\toprule
\multirow{2}{*}{Method} & \multicolumn{2}{c}{\multirow{2}{*}{VeRi}} & \multicolumn{6}{c}{VehicleID}                          & \multicolumn{6}{c}{VERI-WILD}                   \\
                        & \multicolumn{2}{l}{}   & \multicolumn{2}{c}{Small} & \multicolumn{2}{c}{Medium} & \multicolumn{2}{c}{Large} & \multicolumn{2}{c}{Small} & \multicolumn{2}{c}{Medium} & \multicolumn{2}{c}{large} \\ 
          & mAP      & R1        & R1      & R5      & R1     & R5       & R1    & R5    & mAP      & R1      & mAP     & R1       & mAP    & R1 \\ \midrule
VANet \cite{chu2019vehicle}     & 66.3 & 89.8  & 88.1   & 97.3  & 83.2  & 95.1  & 80.4 & 93.0
          &  -- &   -- &   -- &  --  &   --  &   --  \\
PART \cite{he2019part}  &74.3  & 94.3   &  78.4  & 92.3 & 75.0  & 88.3  & 74.2 & 86.4
          & --   & --  &  --   &  --   &  --  &  --  \\
SEVER \cite{khorramshahi2020devil} & 79.6  & 96.4  & 79.9  & 95.2  & 77.6  & 91.1  & 75.3 & 88.3
          & 80.9  & \underline{94.5}    & 75.3   & \underline{92.7}   & 67.7  & \underline{89.5}     \\
SGFD \cite{li2021self}  & 81.0   & \underline{96.7}    & 86.8    & 97.4  & 83.5   & 95.6   & \underline{80.8}  & \underline{93.7}
          & --         &  --         & --        & --        & --      & --     \\
HRC$^{*}$ \cite{zhao2021heterogeneous} & \underline{81.9}  & \textbf{96.8} &87.8 & \underline{98.1}  & 81.7  & \underline{96.3}  & 78.5 & 93.5
          & \underline{85.2}     & 94.0      & \underline{80.0}    & 91.6    & \underline{72.2}   & 88.0     \\
UFDN \cite{qian2022unstructured} & 81.5     & 96.4      & \underline{88.4}   & --  & \underline{84.8}  & --   & 80.6  & --
          & 84.6     & --      & 79.4     & --   & 72.0  & --     \\ \midrule
AdaSP    & \textbf{82.7}   & \underline{96.7}   &  \textbf{89.3}  & \textbf{98.4} & \textbf{85.6}  & \textbf{96.7}     & \textbf{83.0}   & \textbf{94.9}
          & \textbf{89.2}   & \textbf{96.2}   & \textbf{84.8}    & \textbf{95.0}    &\textbf{78.7}  & \textbf{92.1}     \\ \bottomrule       
\end{tabular}
}
\vspace{-2mm}
\end{table*}

\vspace{-3mm}
\paragraph{Impact of hyper-parameters.}
In this section, we explore the impact of hyper-parameters, including the weight $\lambda$ of SP loss in Eq. \ref{eq9} and its temperature $\tau$, on both person and vehicle ReID datasets. We first vary $\lambda$ from $0.01$ to $0.5$ for MSMT17. The experimental results, exhibited in Tab. \ref{tab5}, suggest that the overall performance is not sensitive to the weight $\lambda$ and the mAP can remain above $66.5$ when $\lambda$ falls into $[0.05,0.3]$, while the best performance is achieved when $\lambda$ is set to 0.1. The temperature $\tau$ is then configured from $0.01$ to $0.09$ with an interval of $0.01$ and the experimental results are shown in Tab. \ref{tab6}. It can be seen that the best mAP performance is achieved when the temperature is $0.04$.
We then conduct a similar experiment on the VeRi-776 dataset to explore hyper-parameters' impact on Vehicle ReID tasks. The experimental results are shown in Tab. S3 and Tab. S4 in Supplementary Materials. The best mAP is achieved at a temperature of $0.05$ and a loss weight of $0.5$. Based on the above parameter study, we adopt the optimal hyper-parameters for all experiments in this work.

\subsection{Comparison with State-of-the-art Methods}
We compare AdaSP loss with the state-of-the-art ReID methods based on CNN network architectures on four person ReID datasets and three vehicle ReID benchmarks. The experimental results are shown in Tab. \ref{tab7} and Tab. \ref{tab8}. It can be observed from Tab. \ref{tab7} that AdaSP outperforms MGN and ABD-Net at the same input size of $384\times 128$ by a large margin. Compared to other approaches with the input size of $256\times 128$, AdaSP achieves the best mAP of $64.7$ and $78.0$ on the MSMT17 and CUHK03-D datasets, which is $4.9\%$ and $3.3\%$ higher than the SOTA approach Nformer, respectively. 
For the vehicle ReID benchmarks, our approach achieves the best performance on almost all evaluation metrics across all datasets. Especially for the largest dataset VERI-WILD, AdaSP outperforms HRC\cite{zhao2021heterogeneous} and UFDN \cite{qian2022unstructured} by at least $4\%$ on the mAP performances across the three test sets.

\section{Conclusion}
In this paper, we propose a sparse pairwise loss that only requires excavating one positive and one negative pair for each class in a mini-batch to form metric loss items. Aiming to avoid sampling harmful positive pairs from the identities with large intra-class variations, we develop a least-hard positive mining approach and further endow it with an adaptive strategy.
We implemented a careful evaluation of the proposed SP/AdaSP loss on both person and vehicle ReID datasets. Quantitative experimental results suggest that our loss outperforms current metric losses and achieves state-of-the-art performance on most benchmarks.

\section*{Acknowledgement}
This work was supported by National Natural Science Foundation of China (Grant No.62103228).

{\small
\bibliographystyle{ieee_fullname}
\bibliography{PaperForReview}
}

\newpage
\section*{\centering{Supplementary Materials}}
\setcounter{figure}{0}
\setcounter{table}{0}
\setcounter{equation}{0}
\setcounter{section}{1}

\renewcommand*{\thefigure}{S\arabic{figure}}
\renewcommand*{\thetable}{S\arabic{table}}
\renewcommand*{\theequation}{S\arabic{equation}}

\subsection{Proof}
Unlike traditional dense sampling methods that may suffer from sampling harmful pairs, the proposed Sparse Pairwise (SP) loss 
can effectively avoid introducing harmful positive pairs \textbf{when combined with our least-hard positive mining strategy}. Here, we provide a simple theoretical proof to demonstrate \textbf{the expected percentage of harmful pairs sampled by SP in a mini-batch is lower than that by the traditional dense sampling methods}.

Let us assume that harmful positive pairs always have lower visual similarities than real positive pairs (which is true in general).
Given a mini-batch with $N$ IDs and each ID contains $M$ instances, which shape a positive similarity matrix $\mathbf{S}\in \mathbb{R}^{M\times M}_+$, dense methods sample $M$ positive pairs with the lowest similarities in each row of $\mathbf{S}$. While our approach utilizes the pair with the highest similarity from the $M$ positive pairs sampled by dense methods. 
Suppose there are $K$ harmful positive pairs for each ID, which results in two situations according to the value of $K$:

\textbf{Situation I}: $K<M$. SP does not encounter harmful pairs because at least $M-1$ positive pairs with similarity smaller than the pair sampled by SP, while there are only $K(\leqslant M-1)$ harmful pairs. However, dense methods definitely sample harmful pairs and its expectation of sampling harmful pairs is assumed to be $E^{D}_{I}\in \left( 0, K\right]$. 

\textbf{Situation II}: $K\geqslant M$. SP is possibly sampling harmful pairs and its sampling expectation is assumed to be $E^{SP}_{II}\in [0,1]$ since it only samples one positive pair for each ID. Meanwhile, the expectation of dense methods for sampling harmful pairs is given by: 
\begin{equation}
    E^{D}_{II} = ME^{SP}_{II}+E^{D}_{\ast}\left( 1-E^{SP}_{II} \right)
    \label{eqs1}
\end{equation}
where $E^{D}_{\ast}$ denotes the expectation of dense methods for the ID that SP does not sample a harmful pair.

Assume the numbers of ID with $K<M$ and $K\geqslant M$ in a mini-batch are $U$ and $V$, respectively. The expected percentage of harmful pairs sampled by SP and dense approaches in a mini-batch can be given by: 
\begin{equation}
    P_{SP}=\frac{V E^{SP}_{II}}{N}
\label{eqs2}
\end{equation}

\begin{equation}
\begin{split}
    P_{D}&=\frac{UE^{D}_{I}+V(ME^S_{II}+E^{D}_{\ast}( 1-E^{SP}_{II}))}{NM}
    \\
    &=\frac{UE_{I}^{D}}{NM}+\frac{VE_{II}^{SP}}{N}+\frac{VE_{\ast}^{D}\left( 1-E_{II}^{SP} \right)}{NM}
\end{split}
\label{eqs3}
\end{equation}
Therefore, $P_{SP}<P_D$ is concluded. Especially for the ID with $K<M$, SP can completely avoid sampling harmful positive pairs.

\subsection{Additional Results}

\begin{table}
  \centering
  \footnotesize
  \caption{Statistics of object ReID datasets utilized in this work.}
  \label{tabs1}
   \resizebox{\linewidth}{!}{
  \begin{tabular}{lcc|lcc}
  \toprule
   Person  &\#ID  &\#images  &Vehicle  &\#ID  &\#images   \\ \midrule
      CUHK03 &1467 &13164  &VeRi-776 &776 &49357\\
                            Market-1501 &1501 &32668 &VehicleID   &26328 &221567\\
                            DukeMTMC   &1404 &36411 &VERI-WILD   &40671 &416314\\
                            MSMT17   &4101 &126441 & -- & -- &--\\ 
  \bottomrule
  \end{tabular}
  }
\end{table}

\begin{figure}
    \centering
    \includegraphics[scale = 0.6]{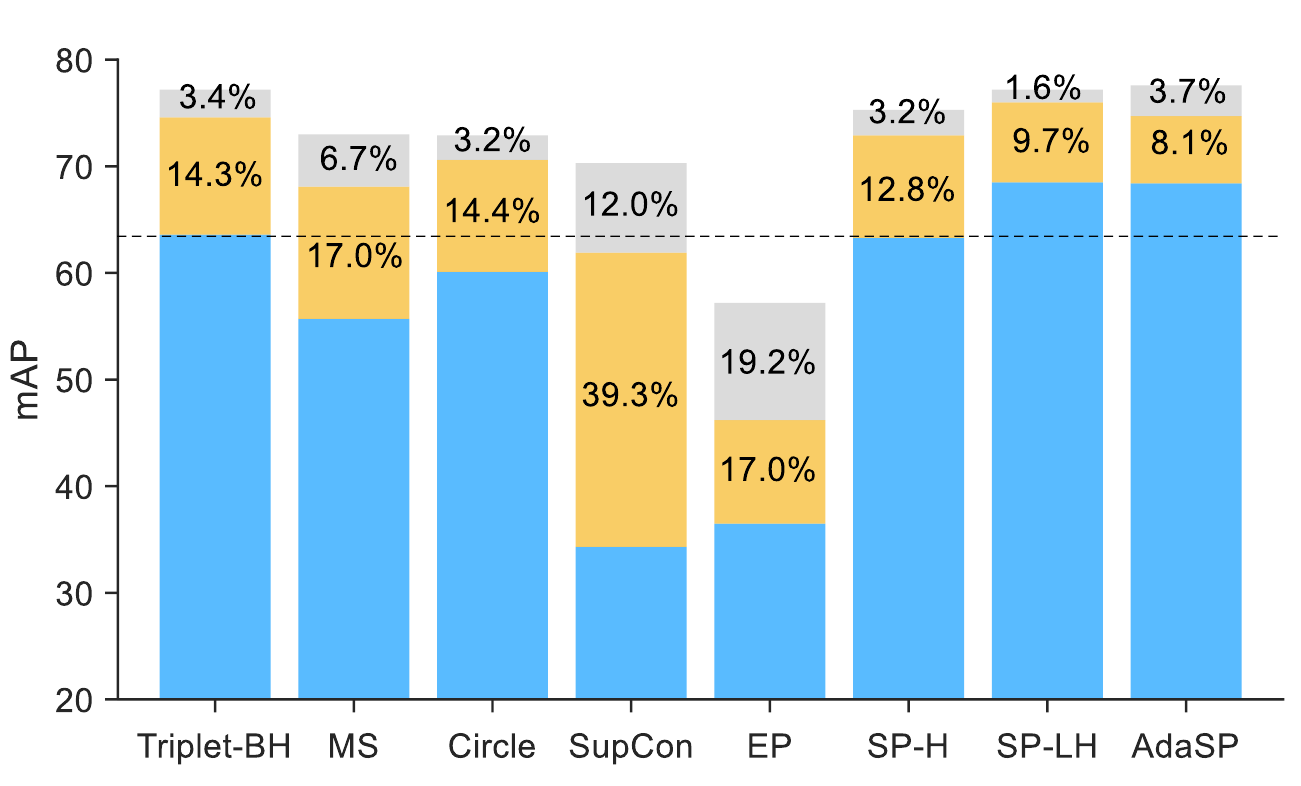}
    \caption{Comparison on the robustness of the performance regarding the number of instances per identity (including 8, 16, and 32) in a mini-batch on VeRi-776 dataset. The numbers in each bar represent the percentage of dropped performance relative to the best performance achieved in 8 instances. The dashed line denotes the mAP performance of SP-H with 32 instances per identity.}
    \label{figs1}
\end{figure}

\paragraph{Robustness to increasing intra-class variations.}
To demonstrate the robustness of SP loss towards increasing intra-class variations, we also evaluate various metric losses on the VeRi-776 dataset. The experimental results are exhibited in Fig. \ref{figs1}. Consistent with the results on the MSMT17 dataset, shown in the main text, all metric losses still suffer from a performance decline as the intra-class variations increase. It can be observed that the EP loss achieves very limited mAP even though $8$ instances per identity are utilized. Besides, the performance of SupCon loss drops sharply and the MS loss loses more than $23\%$ of mAP with a growing number of instances. The triplet loss still outperforms other dense pairwise losses and achieves comparable performance with our SP-H loss. However, it can be seen that SP loss variants with the least-hard and the adaptive positive mining strategies lose less than $12\%$ of mAP when the instance number increases to $32$, indicating considerable robustness toward increasing intra-class variations.

\begin{table*}
\centering
\small
\caption{Experimental results on person ReID datasets using the backbones of ResNet50-IBN, ResNet-152 and MGN with the input size of $256 \times 128$. The loss function consists of a cross-entropy loss and a metric loss (AdaSP for our approach, Triplet-BH for comparison).}
\begin{tabular}{ll|cc|cc|cc|cc|cc}
\toprule
\label{tabs2}
 \multirow{2}{*}{Backbone} & \multirow{2}{*}{Loss} & \multicolumn{2}{c}{MSMT17} & \multicolumn{2}{c}{Market1501} & \multicolumn{2}{c}{DukeMTMC} & \multicolumn{2}{c}{CUHK03-L} & \multicolumn{2}{c}{CUHK03-D} \\  
   &   & mAP   & R1    & mAP  & R1      & mAP    & R1  &mAP &R1 &mAP &R1  \\ \midrule

\multirow{2}{*}{ResNet50-IBN} &Triplet-BH  & 57.3  & 80.0   & 83.3 & 93.1 & 75.0  & 86.6 & 65.5 & 67.6 & 62.1 & 65.0  \\
&AdaSP       & \textbf{61.7}  & \textbf{83.7}  & \textbf{87.7} & \textbf{95.4}  & \textbf{79.8}  & \textbf{90.3} & \textbf{69.7} & \textbf{71.4} & \textbf{67.6} & \textbf{70.3} \\ \midrule

\multirow{2}{*}{ResNet-152} &Triplet-BH  &57.9	&80.2	&84.8	&93.7	&77.1	&88.1	&69.3	&72.1	&64.8	&68.0  \\

&AdaSP       & \textbf{60.1}  & \textbf{82.4}  & \textbf{88.1} & \textbf{95.0}  & \textbf{80.0}  & \textbf{89.4} & \textbf{71.6} & \textbf{73.2} & \textbf{69.2} & \textbf{79.4} \\  \midrule

\multirow{2}{*}{MGN} &Triplet-BH  & 59.2  & 80.8   & 88.1 & 95.0 & 79.1  & 88.6 & 74.9 & 76.9 & 72.2 & 75.1  \\
&AdaSP       & \textbf{60.6}  & \textbf{82.1}  & \textbf{88.5} & \textbf{95.5}  & \textbf{80.3}  & \textbf{90.1} & \textbf{77.7} & \textbf{79.7} & \textbf{74.5} & \textbf{77.3} \\

\bottomrule
\end{tabular}
\end{table*}

\paragraph{Generalizing to different networks.}
We also test SP loss on other networks, including ResNet50-IBN, ResNet-152 and MGN, with the input size of $256\times128$. The experimental results are shown in Tab. \ref{tabs2}. Compared to ResNet-50, the performance of Triplet-BH on each dataset is significantly increased by these well-designed networks. Nevertheless, our approach AdaSP can still facilitate the ReID performance on each dataset, especially the hard ones, such as MSMT17 and CUHK03. 

\balance

\paragraph{Impact of hyper-parameters.}
We adopt the VeRi-776 dataset to explore the impact of hyper-parameters on Vehicle ReID tasks. The experimental results are exhibited in Tab. \ref{tabs3} and \ref{tabs4}. The best mAP is achieved at a temperature of $0.05$ and a weight of $0.5$. In addition, it can be seen that all the mAPs are higher than $82.0$ when the weight of SP loss varies from $0.1$ to $0.9$, suggesting that the vehicle ReID performance is not sensitive to the weight of SP loss.

\begin{table}
\centering
\footnotesize
\caption{Experimental results of the ablation study on the weight for metric loss $\lambda$ in VeRi-776 dataset.}
\label{tabs3}
\begin{tabular}{lccccccc}
\toprule
$\lambda$   & 0.1   &0.3    &0.5        & 0.7       & 0.9 \\ \midrule
mAP       & 82.0     & 82.5   & \textbf{82.7}   & 82.4   & 82.2     \\
R1        & \textbf{97.1}     & 97.0   & 96.7   & 96.7   & 96.5   \\   \bottomrule   
\end{tabular}
\end{table}

\begin{table}
\centering
\caption{Experimental results of the ablation study on the temperature $\tau$ in VeRi-776 dataset.}
\label{tabs4}
\resizebox{\linewidth}{!}{
\begin{tabular}{lccccccccc}
\toprule
$\tau$ &0.01   & 0.02 & 0.03 &0.04  &0.05  &0.06  & 0.07  & 0.08  &0.09 \\ \midrule
mAP     &67.6  & 76.0 & 79.9  & 82.6 & \textbf{82.7}  & 82.1 & 81.6 & 78.9 & 77.1  \\
1      &90.5  & 94.3 & 95.5  & \textbf{96.8} & 96.7  & 97.0 & 96.4 & 96.0 & 95.1   \\   \bottomrule   
\end{tabular}
}
\end{table}

\end{document}